\pdfoutput=1
\documentclass[11pt]{article}

\usepackage{emnlp2021}

\usepackage{times}
\usepackage{latexsym}

\usepackage[T1]{fontenc}
\usepackage[utf8]{inputenc}

\usepackage{microtype}
\usepackage{rotating}
\usepackage{multirow}
\usepackage{booktabs}
\usepackage{pifont}
\usepackage{amsmath}
\usepackage{subcaption}

\newcommand{\bftab}{\fontseries{b}\selectfont}
\title{Multi-modal Retrieval of Tables and Texts Using Tri-encoder Models}

\author{Bogdan Kostić \and Julian Risch \and Timo Möller \\
deepset\\ \texttt{\{bogdan.kostic, julian.risch, timo.moeller\}@deepset.ai}}

\begin{document}
\maketitle
\begin{abstract}
Open-domain extractive question answering works well on textual data by first retrieving candidate texts and then extracting the answer from those candidates. However, some questions cannot be answered by text alone but require information stored in tables.
In this paper, we present an approach for retrieving both texts and tables relevant to a question by jointly encoding texts, tables and questions into a single vector space.
To this end, we create a new multi-modal dataset based on text and table datasets from related work and compare the retrieval performance of different encoding schemata.
We find that dense vector embeddings of transformer models outperform sparse embeddings on four out of six evaluation datasets. 
Comparing different dense embedding models, tri-encoders with one encoder for each question, text and table increase retrieval performance compared to bi-encoders with one encoder for the question and one for both text and tables. We release the newly created multi-modal dataset to the community so that it can be used for training and evaluation.
\end{abstract}

\section{Introduction}
\label{sec:introduction}
Finding the answer to a factual question in a large collection of documents is a tedious task that many people from a broad range of domains have to complete on a daily basis. 
In order to address this task with machine learning approaches, it has been formalized as open-domain extractive question-answering (QA). 
More specifically, given a natural language question and a database of text documents as a knowledge base, open-domain extractive QA aims to extract a substring that answers the given question out of one of the documents.

The standard approach for this task is a pipeline architecture consisting of two components: a retriever selecting a small subset of relevant documents from the database and a reader extracting granular answers out of each of these retrieved documents~\cite{voorhees1999trec}. 
%This paper focuses on the retriever, which takes the natural language question as input and returns a small subset out of the document database that might contain the answer to the question. 
In this paper, we focus on the retriever and present a transformer-based tri-encoder model as an implementation of this component.
Retrievers can also be implemented with bag-of-words retrieval methods, such as TF-IDF or BM25, that transform all documents and the question to sparse vector representations. 
However, as these methods rely on a lexical overlap of the question and the documents, they fail to capture synonymy and other semantic relationships.
This limitation motivates the use of dense vector representations and we compare dense retrieval models to a BM25 baseline in our experiments.
A survey on term-based, early semantic, and neural semantic models for document retrieval as a first step before document reranking and down-stream tasks, such as question answering, has been published by \citet{cai2021semantic}.
%, for example, Dense Passage Retrieval by \citet{Karpukhin.2020}. relying on two distinct transformer-based language models, tries to solve this problem by embedding both the question and the text documents into a dense vector space. This approach outperforms BM25 by a large margin improving the performance of the whole open-domain QA pipeline as well.

To date, most of the research centered around question-answering focuses on using free-form text as the single source for answering questions. 
However, valuable information can obviously be found in other modalities as well. 
For instance, a lot of information is stored in semi-structured tables; according to \citet{Cafarella.2008}, more than 14.1 billion tables can be found on the World Wide Web. %
%Nevertheless, not much research has been conducted on successfully using this source of information to answer factual questions. 
%
%This results in having two separate pipelines for answering factual questions: one pipeline using free-form texts as a knowledge base and the second pipeline using tables.
%
%However, 
Given that a user typically does not know in advance in which modality the answer to their question resides, a QA system capable of jointly handling text and tables is needed. 
One major challenge in building such a system is to represent texts and tables in a way that allows capturing semantic similarity and retrieving texts and tables that are semantically related to a given question.

\paragraph{Contributions} The contributions of this paper can be summarized as follows:
(1) we present bi-encoder and tri-encoder models that are capable of joint retrieval of tables and texts;
(2) we create and release a multi-modal dataset for training and evaluating models on this task;\footnote{\url{https://multimodalretrieval.s3.eu-central-1.amazonaws.com/data.zip}}
(3) we compare sparse retrieval models with dense retrieval models using bi-encoders and tri-encoders on our new multi-modal dataset and on five uni-modal datasets from related work.

\paragraph{Outline} The remainder of this paper is structured as follows: Section~\ref{sec:related_work} summarizes existing methods for uni-modal retrieval of texts on the one hand and of tables on the other hand. Further, it discusses the only two, recently published approaches for joint retrieval of tables and texts.
Section~\ref{sec:datasets} briefly describes the existing uni-modal datasets and our new multi-modal dataset, which we use to train the retrieval models presented in Section~\ref{sec:approach} and to evaluate these models in Section~\ref{sec:experiments}.
Section~\ref{sec:conclusion} concludes the paper and gives directions for future work. 

\section{Related Work}
\label{sec:related_work}
%Todo: add three new papers from ACL 2021 \cite{zhu-etal-2021-tat} and \cite{pan-etal-2021-cltr} and \cite{li-etal-2021-dual}
To the best of our knowledge the only existing work that addresses joint retrieval of tables and texts is by \citet{Chen.2021}, \citet{talmor2021multimodalqa} and \citet{li-etal-2021-dual}. 
%They split tables into segments and text documents into passages to obtain smaller basic retrieval blocks. 
To address the challenge of limited context available for tables, \citet{Chen.2021} fuse a table segment and text passages into one block if they mention the same named entities.
Each block is represented with a single dense embedding so that a table and relevant passages are jointly retrieved as a group if the embedding is similar to that of the question. 
This grouping makes sense because \citet{Chen.2021} address the task of multi-hop QA, where information needs to be aggregated from multiple tables and texts to answer a question.
In contrast to that, we address the slightly different task of single-hop QA, where either only one table or one text is needed to answer a question.
Therefore, our approach represents tables and texts with separate embeddings, which are in the same embedding space.
The advantage here is that the model learns to estimate relevance on the more fine-grained level of individual tables or texts and can decide whether a particular table or text is more relevant to the question.

\citet{li-etal-2021-dual} also address the task of multi-hop QA on texts and tables but retrieve them individually. They make use of the sparse retrieval method BM25 and a transformer-based reranker to generate a set of candidate texts and tables. With two separate BM25 indices for texts and tables, they retrieve a set of documents for each modality. In a second step, they apply a joint BERT-based reranker to reduce the size of candidate texts and tables.
\citet{talmor2021multimodalqa} create a \textsc{MultiModalQA} dataset containing questions that require joint reasoning over tables, texts, and images. 

Otherwise closely related to our approach are \textsc{TaBERT} by \citet{yin-etal-2020-tabert} and \textsc{TaPas} by \citet{Herzig.2020}, who use tables and texts for language model pre-training but do not consider joint retrieval of tables and texts.
\textsc{TURL} focuses on representation learning for tables but also does not consider the retrieval task~\cite{10.5555/3430915.3442430}.
Due to this limited amount of prior research, we discuss related work on the separate tasks of uni-modal text retrieval and table retrieval in the following.

\subsection{Text Retrieval}
Dense Passage Retrieval (DPR) by \citet{Karpukhin.2020} relies on a bi-encoder model comprising two separate BERT models \cite{Devlin.2019}.
Similar to TF-IDF, DPR is a vector space model that represents both queries and documents in the same vector space. 
However, while TF-IDF represents text documents as very high dimensional sparse vectors, DPR relies on relatively low dimensional dense embeddings. 
%Given that BERT restricts the maximum input sequence length to 512 tokens, longer text documents are split into smaller passages that can be fed as input to a BERT model. 
While one of the models, the passage encoder (BERT$_p$), is used to encode text passages at indexing time, the second model, the question encoder (BERT$_q$), is used to encode questions at query time. 
Since BERT's [CLS]-token is particularly designated to capture the meaning of the whole input sequence, its embedding is used as a representation vector for both the text passages and the questions. 
%Query vectors $\vv{q}$ for queries $q$ and passage vectors $\vv{p}$ for passages $p$ are therefore computed as: 
%\begin{equation}
%    \vv{q} = \text{BERT}_q(q)[\text{CLS}]\\
%\end{equation}
%\begin{equation}
%    \vv{p} = \text{BERT}_p(p)[\text{CLS}]
%\end{equation}

%Given that DPR is a vector space model, the text passages are at retrieval time ranked with regard to the similarity of their embeddings to the embedding of the question using the dot product or cosine similarity. 
%The goal of the training is, therefore, to build an embedding model that embeds questions and their relevant passages close to each other. 
The training aims to increase the dot product or cosine similarity of semantically similar passages and questions. 
In order to achieve this, \citet{Karpukhin.2020} use, besides the question's positive passage, hard negative passages as well as in-batch negative passages as training signal. Hard negatives are sampled utilizing a BM25-based retriever on the whole English Wikipedia dump. For each question, they use the highest ranked passage not containing the question's answer string. 
%Using in-batch negatives makes the training computationally a lot more efficient. Positive and hard-negative contexts for a given query are treated as negative contexts for the other queries in the same batch. This allows to compute the embedding of each passage in the batch only once while using it with all query embeddings in that batch.
%
%\citet{Karpukhin.2020} showed that their approach using dense embeddings
DPR drastically outperforms BM25 by almost 20 percentage points with regard to recall@20 on the Natural Questions (\textsc{NQ}) dataset by \citet{Kwiatkowski.2019}. 
%This results in an overall improvement of the whole open-domain QA pipeline.

\subsection{Table Retrieval}
%The few papers that have been published on this topic were mostly concerned with a closed-domain setting where the table containing the answer to the question is already given, i.e., no retrieval component is needed. 
%The classical approach has been to use sequence-to-sequence models to transform the natural language question into a logical representation that can be used to query the table. 
%The release of \textsc{TaPas} \cite{Herzig.2020, Eisenschlos.2020}, a transformer-based language model capable of capturing the two-dimensional structure of tables, set a new state-of-the-art for closed-domain QA on tables without the need of creating this intermediate representation.
%
%Since this table reader component is not enough for performing efficiently open-domain QA on a large database of distinct tables, the logical next step is to ascertain a good enough retriever to be able to adapt the classical QA pipeline to tables. 
%With Dense Table Retrieval, \citet{Herzig.2021} confirmed that the same is valid for tables as for free-form text: dense vector embeddings can be used for retrieving tables and perform much better than sparse vector methods like BM25. 
%
%

Most existing table retrieval approaches rely on supervised learning-to-rank approaches. 
\citet{Zhang.2018} combine a set of hand-crafted query features, table features and query-table features with semantic similarity of table and query as additional feature. 
To get table and query representations, they use the average of pre-trained word2vec~\cite{Mikolov.2013} and RDF2vec embeddings~\cite{Ristoski.2016}. 
These features are then used to train a random forest regressor to get relevance scores of the tables with regard to the query.
%\citet{Zhang.2019} try to further improve this approach by using word embeddings that are trained on a corpus of tables. 
%This, however, does not give a performance boost compared to general word embeddings trained on a corpus of text documents.
Interestingly, training word embeddings on a corpus of tables instead of texts does not improve performance~\cite{Zhang.2019}.

\citet{Shraga.2020a} combine intrinsic and extrinsic table similarity scores. 
For the intrinsic table similarity, they concatenate the table's title, caption, header and data rows and use a sliding window over this concatenated string to get a set of candidate passages for each table. 
Each candidate passage is scored using BM25 and the maximum of all passages of a table is the intrinsic score with regard to a query.
%Under the assumption that similar tables behave similarly when it comes to the relevance of a table with regard to a user's information needs, 
The extrinsic score is based on table-table similarities to ensure that similar tables get a similar final score.

\citet{Bagheri.2020} focus on hard queries that contain terms that do not occur in the relevant tables, which means the query and the relevant tables have a low lexical overlap.
To this end, they learn low dimensional latent factor matrices to represent tables as well as queries, i.e., they learn term co-occurrences to be able to get tables that address the same topic but only partially overlap on a lexical level.
Based on this result, we compare results on datasets with high or low lexical overlap in our experiments.

There are four deep learning approaches for table retrieval~\cite{Shraga.2020b,pan-etal-2021-cltr,Chen.2020,Herzig.2021}.
\citet{Shraga.2020b} treat tables and queries as multi-modal objects that consist of query, table caption, schema, rows and columns. 
Each component is encoded using its own neural network encoder that accounts for its special characteristics. 
%Given that both the table caption and the query are composed of natural language, \citet{Shraga.2020b} use recurrent neural networks to represent them as vectors. 
%Since different orders inside a table schema do not make a difference, table schemas are encoded utilizing an order-agnostic multi-layer perceptron. Ultimately, to encode both the table rows and columns, they make use of convolutional neural networks to capture the spatial relationships of cells close to each other.
Subsequently, these uni-modal encodings are joined into a single representation, which is passed on to fully connected layers that predict whether the input table is relevant with regard to the input query.

\citet{pan-etal-2021-cltr} stack two retrieval components. First, they use BM25 to produce a large subset of possibly relevant tables. These tables are then passed to a row-column intersection model that generates a probability distribution over the table cells whether they contain the answer to the user's query. The maximum cell-level score for each table represents its retrieval score.

\citet{Chen.2020} apply the transformer-based language model BERT \cite{Devlin.2019} to the table retrieval task by combining BERT embeddings with other, hand-curated table and query features. 
Their approach consists of several components, including a step where the concatenation of a query, a table's context fields and selected relevant table rows are processed by a BERT model.
%The content selector first selects relevant rows inside the table for the query by applying cosine similarity on the word embeddings of the query and the word embeddings of the table. 
%Next, they concatenate the query, the table's context fields and the selected relevant table rows and use this string as input to a BERT network. 
%Furthermore, they send the manually curated features from the query, the table and the table's context fields through a fully connected layer. 
%Both outputs are then concatenated to a single vector which is used as input to a final regression layer returning a relevance score of the table with regard to the query.
A significant downside of this approach is that it becomes inefficient with increasing number of tables: 
for each query, all tables need to be passed through a BERT network. 
%This is computationally quite costly and makes the need for a retriever component inside the open-domain QA pipeline obsolete, since it would not reduce computation time compared to just passing all the tables to the reader component. 
%This table retrieval approach would, therefore, not be suitable as a retriever component inside an open-domain QA pipeline.

\citet{Herzig.2021} solve this efficiency problem by adapting \citeauthor{Karpukhin.2020}'s (\citeyear{Karpukhin.2020}) dense passage retrieval approach to dense table retrieval (DTR). 
To this end, they make use of \textsc{TaPas}~\cite{Herzig.2020}, a transformer-based language model that has been pre-trained on millions of tables. 
\textsc{TaPas} extends BERT by adding three different types of positional embeddings to encode the two-dimensional tabular structure: row, column and rank embeddings. 
This allows to flatten the table by concatenating the rows to a one-dimensional sequence of tokens.
%without losing any information deriving from the two-dimensional structure of the table. 
%Pre-training of \textsc{TaPas} was done similarly to the pre-training of BERT using the masked language modeling objective on a large number of tables from Wikipedia along with table caption, article title, article description, segment title, and text of the segment the table occurs in.
%
Similar to \citet{Karpukhin.2020}, \citet{Herzig.2021} make use of a bi-encoder approach. 
However, they use two \textsc{TaPas} instances instead of BERT instances to encode the queries and the tables, respectively. 
%DTR is, therefore, a vector space model specifically designed for the retrieval of tables. 
The goal of training this bi-encoder is to build an embedding model that generates similar embeddings for questions and their relevant tables. 
As in \citeauthor{Karpukhin.2020}'s (\citeyear{Karpukhin.2020}) approach, this goal is achieved using hard-negatives retrieved from all the tables from the English Wikipedia dump as well as in-batch negatives. 
DTR outperforms BM25 by more than 40 percentage points on the \textsc{NQ-Tables} dataset \cite{Herzig.2021}. 
However, the experiments also show that \textsc{TaPas} requires additional pre-training on the task of table retrieval on millions of tables scraped from Wikipedia. 
As a further research direction for future work, \citet{Herzig.2021} propose to combine tables and texts for multi-modal open-domain QA. 
We contribute towards this goal in our paper by providing a multi-modal retriever as one component of a multi-modal open-domain QA pipeline on tables and texts.

%While transformer-based models started as a technique to apply transfer learning on language-specific tasks, more and more models are released that adapt the attention mechanism characteristic for transformer-based models to non-textual data. 
%To date, there also exist transformer-based models for images \cite{Tan.2019, Li.2020, Dosovitskiy.2021, Li.2021, Radford.2021, Touvron.2021}, videos \cite{Bertasius.2021} and speech \cite{Baevski.2020, Hsu.2021}.

\section{Datasets}
\label{sec:datasets}
The training and evaluation of models examined in this paper on the task of multi-modal retrieval makes use of five datasets from related work: \textsc{NQ} \cite{Kwiatkowski.2019}, \textsc{NQ-Tables} \cite{Herzig.2021}, \textsc{WikiSQL} \cite{Zhong.2017}, a subset of \textsc{WikiSQL}, which we call \textsc{WikiSQL}$_{\text{ctx-independent}}$, and \textsc{OTT-QA} \cite{Chen.2021}. This section briefly explains the characteristics of these datasets and of our newly created multi-modal retrieval dataset comprising tables and texts, which we call \textsc{MultiModalRetrieval}. 
Table~\ref{table:Datasets} gives an overview of the modality and the number of samples in each dataset.

\begin{table*}[th]
\centering
\begin{tabular}{lrrrc}
    \toprule
     \textbf{Dataset} &  \textbf{Modality}& \textbf{Train} & \textbf{Test} & \textbf{Ctx-ind.} \\
     \midrule
     \textsc{NQ} & text & 58,880 & 3,610 & \ding{52} \\
     \textsc{NQ-Tables} & table & 9,594 & 966 & \ding{52} \\
     \textsc{WikiSQL} & table & 56,355 & 15,878 & \ding{54} \\
     \textsc{WikiSQL}$_{\text{ctx-independent}}$  & table & 7,336 & 2,101 & \ding{52} \\
     \textsc{OTT-QA} & table & 41,469 & 2,158 & \ding{52}\\
     \textsc{MultiModalRetrieval} & text \& table & 120,239 & 4,937 & \ding{52} \\
    \bottomrule
\end{tabular}
\caption{Modality and number of train and test samples for multi-modal retrieval models. Only \textsc{WikiSQL} is not context-independent (ctx-ind.), which is why we create the subset \textsc{WikiSQL}$_{\text{ctx-independent}}$.}
\label{table:Datasets}
\end{table*}
%\textsc{OTT-QA} table\footnote{Most questions are answered by both text and table, but only the gold tables are given.}

\paragraph{NQ}
Natural Questions (\textsc{NQ}) \cite{Kwiatkowski.2019} is an open-domain QA dataset on Wikipedia articles. It consists of questions, their answers and the text passages the answers reside in. The questions are \emph{natural} because they consist of real user queries issued to the Google search engine instead of questions posed by annotators after reading a text passage, which was done to create other popular QA datasets, such as the Stanford Question Answering Dataset (SQuAD) \cite{Rajpurkar.2018}. Having \emph{natural} questions does not only ensure that the questions correspond to information needs by real users but also makes the dataset open-domain, i.e., the questions are context-independent and can be understood without their accompanying text passage that contains the answer.
For the purpose of retrieval, we utilize \citeauthor{Karpukhin.2020}'s (\citeyear{Karpukhin.2020}) preprocessed variant of \textsc{NQ}. 
%It consists of 58,880~questions in the training set, 8,757 questions in the development set and 3,610 questions in the test set.

\paragraph{\textsc{NQ-Tables}}
The answers to most of the questions inside the \textsc{NQ} dataset can be found in plain unstructured text. However, a subset of the questions are answered by tables. As a consequence, \citet{Herzig.2021} construct \textsc{NQ-Tables}, a table-specific QA dataset based on \textsc{NQ}. To achieve this objective, they extract all the tables and all the questions whose answers reside inside a table. They come up with a dataset consisting of 9,594 questions in the training set, 1,068 questions in the development set, 966 questions in the test set and 169,898 tables in total. Given that this dataset is a subset of \textsc{NQ}, the questions share the characteristic of being context-independent.

\paragraph{\textsc{WikiSQL}}
\label{sec:WikiSQL}
\textsc{WikiSQL} \cite{Zhong.2017} is a closed-domain QA dataset consisting of 24,241 tables and 80,654~natural language questions together with their corresponding SQL query. To build this dataset, \citet{Zhong.2017} generate a number of random SQL queries for each table. These SQL queries are then transformed into crude questions using templates. Finally, Amazon Mechanical Turk crowd workers paraphrase these crude questions into natural language questions, which are checked by two additional crowd workers.

\paragraph{\textsc{WikiSQL}$_{\text{ctx-independent}}$}
Since \textsc{WikiSQL} is a closed-domain QA dataset, the majority of its questions is context-dependent, i.e., they do not provide enough context to be answered without their accompanying table and are therefore not suitable for training or evaluation on the retrieval task. One example for such insufficient questions inside the dataset is: \emph{``Who is the player that wears number 42?''}, which cannot be answered without additional context given in the table, such as the name of a sports team and a year.
As a consequence, to make use of \textsc{WikiSQL}, all questions that do not provide enough context for retrieval need to be filtered out. For automating this filtering, we labeled a subset of \textsc{WikiSQL}'s questions with regard to whether they are either context-independent or under-specified resulting in 4,553 labels as training set and 612 labels as test set. These labels are then used to train a classifier that predicts whether a question provides enough context. 
%The Framework for Adapting Representation Models (FARM)\footnote{\url{https://github.com/deepset-ai/FARM}} is used to fine-tune a RoBERTa-base \cite{Liu.2019} language model achieving an accuracy of 0.8134 and a macro-averaged F1-score of 0.7748 on the test set. 
We fine-tune a RoBERTa-base \cite{Liu.2019} language model achieving an accuracy of 0.8134 and a macro-averaged F1-score of 0.7748 on the test set. 
Next, we apply this classifier to the whole \textsc{WikiSQL} dataset to filter out all the questions that are predicted as under-specified.
%, resulting in 7,336 questions in the train set, 2,101 questions in the test set and 1,187 questions in the development set.

\paragraph{\textsc{OTT-QA}}
\textsc{OTT-QA} \cite{Chen.2021} is an open-domain multi-hop QA dataset of texts and tables from Wikipedia built upon HybridQA \cite{Chen.2020hybrid}, a closed-domain multi-hop QA dataset. Multi-hop QA refers to the fact that most questions inside the dataset require a combination of different texts and/or tables instead of a single document in order to be answered. To generate an open-domain version of HybridQA, \citet{Chen.2021} let crowd workers decontextualize all the questions. Furthermore, they add additional question-answer pairs on newly crawled tables. 
%The final dataset consists of 41,469 questions in the training set, 2,214 questions in the development set, and 2,158 questions in the test set. Besides, the dataset comprises a total of 419,183~tables.
Since the published annotations contain only the gold tables but not the gold texts, we use \textsc{OTT-QA} only for generating table retrieval training samples and evaluating the uni-modal retrieval of tables.

\paragraph{\textsc{MultiModalRetrieval}}
\label{section:MMDataset}
%Since the goal of this paper is to build a model that is able to retrieve both texts and tables, a retrieval dataset that combines both modalities is needed. 
Given that no multi-modal dataset of tables and texts is readily available, this paper newly introduces such a dataset based on datasets from related work. For this purpose, we combine the question-passage pairs from \textsc{NQ} as questions requiring a text passage to be answered with the question-table pairs from \textsc{NQ-Tables}, \textsc{WikiSQL}$_{\text{ctx-independent}}$, and \textsc{OTT-QA} as questions requiring a table to be answered. Both \citet{Karpukhin.2020} for text retrieval and \citet{Herzig.2021} for table retrieval show that adding hard negatives as training signal boosts the retrieval performance of the model significantly. Therefore, we index 21 million Wikipedia passages (from \citet{Karpukhin.2020}) and 7 million Wikipedia tables (used by \citet{Eisenschlos.2020} to pre-train \textsc{TaPas}) with Elasticsearch to sample hard negatives using BM25. For each question, the highest ranked passages or tables that do no contain the answer string were chosen, i.e., a question originating from a tabular question-answering dataset can also have a text passage as hard negative, and vice versa. %The final training set consists of 120,239 questions. The evaluation set uses a random sample of questions of each dataset, resulting in 4,937 questions.

\section{Approach}
\label{sec:approach}
Our two approaches for the joint retrieval of tables and texts comprise bi-encoder and tri-encoder models based on uni-modal dense retrieval methods by \citet{Karpukhin.2020} and \citet{Herzig.2021}.
%To achieve the goal of creating a joint embedding space for the retrieval of both text and tables, we extend \citeauthor{Karpukhin.2020}'s (\citeyear{Karpukhin.2020}) and \citeauthor{Herzig.2021}'s (\citeyear{Herzig.2021}) uni-modal dense retrieval methods for text or tables, respectively, in two different ways.
%
%Our first approach keeps \citeauthor{Karpukhin.2020}'s (\citeyear{Karpukhin.2020}) bi-encoder architecture. 

In our first approach, the bi-encoder uses one language model to encode the questions and a second model to encode both the tables and the text passages.
Our second approach adds a third encoder, such that there is one separate encoder for questions, text passages and tables. 
%The advantage of this architecture, which we call tri-encoder, is that 
%These three encoders are jointly trained, just as for bi-encoder models. 
In contrast to the bi-encoder where tables and texts are encoded by the same model, the tri-encoder routes tables to the table encoder model and text passages to the text encoder model.

%\section{Experiments}
%\label{sec:experiments}

We train three different bi-encoders and three different tri-encoders, which differ in the underlying language models as specified in Table~\ref{table:encoders}.
%\subsection{Bi-Encoders}
%In total, we train three different bi-encoders, as specified in Table~\ref{table:bi-encoders}, that differ in the underlying language models that are used as encoders. 
The first multi-modal bi-encoder consists of two different BERT-small instances that serve as question encoder and table and text encoder, respectively. Given that BERT models only allow one-dimensional strings of text as input, the two-dimensional tables are transformed into one dimension by concatenating the titles of the page and the section the table occurs in, the caption of the table, and each row of the table.  In order to analyze whether the table-specific language model \textsc{TaPas} \cite{Herzig.2020, Eisenschlos.2020} gives a performance boost compared to a plain BERT model, the remaining two bi-encoders make use of a \textsc{TaPas} model that is pre-trained for the task of table retrieval \cite{Herzig.2021} for at least one of their encoders. Thus, the second bi-encoder uses a BERT-small instance as question encoder and a \textsc{TaPas}-small instance as table and text encoder. The third bi-encoder utilizes two \textsc{TaPas}-small models, one to encode the questions and the second to encode both text and tables.
Using BERT-small instead of BERT-base or BERT-large models drastically reduces the number of parameters and allows to fit more training samples into one batch.
In contrast to a BERT-large model with 24 transformer layers, hidden representations of size 1024, 16 attention heads, and a total number of 335M parameters, BERT-small consists of only 4 transformer layers, hidden representations of size 512, 8 attention heads, and a total number of 29.1M parameters.
%large: 24-layer, 1024-hidden, 16-heads, 335M parameters.
%base: 12-layer, 768-hidden, 12-heads, 110M parameters.
%small: 4-layer, 512-hidden, 8-heads, 29.1M parameters
%number of transformer layers, size of hidden representation, number of attention heads

%\subsection{Tri-Encoders}
%We compare three distinct tri-encoders, as specified in Table~\ref{table:tri-encoders}, to each other that differ in the underlying language models that serve as encoders. 
The first tri-encoder model uses BERT-small instances for all of its three encoders. Also for the tri-encoder approach, we analyze the impact of using \textsc{TaPas} for at least one of the encoders. Therefore, the second architecture makes use of a BERT-small model for both the question encoder and the text encoder and utilizes a \textsc{TaPas}-small instance to encode the tables. The third tri-encoder model uses two \textsc{TaPas}-small instances to encode questions and tables and a BERT-small instance to encode text passages.

\citet{Herzig.2021} use an additional down-projection layer to reduce the dimensionality of the question and table embeddings. We evaluated models with and without such a down-projection layer and found that their results do not differ significantly. Given that the models using an additional down-projection layer are more complex than models that directly utilize the embedding of the [CLS]-token, we consider only \textsc{TaPas}-models without a down-projection layer throughout the remainder of this paper, including the experiments.

\begin{table}
\setlength{\tabcolsep}{4pt}
\renewcommand{\arraystretch}{1.1}
\centering
\begin{tabular}{lccc}
    \toprule
    & \multicolumn{3}{c}{\textbf{Encoders}}\\
    &Question & Text & Table\\
    \midrule
    \multirow{3}{*}{\begin{turn}{90}{\footnotesize Bi-encoder\ \ }\end{turn}} & BERT-small & \multicolumn{2}{c}{------ BERT-small ------}\\
    &BERT-small & \multicolumn{2}{c}{------ \textsc{TaPas}-small ------}\\
    &\textsc{TaPas}-small & \multicolumn{2}{c}{------ \textsc{TaPas}-small ------}\\
    \midrule
    \multirow{3}{*}{\begin{turn}{90}{\footnotesize Tri-encoder\ \ }\end{turn}} &
    BERT-small & BERT-small & BERT-small\\
    &BERT-small & BERT-small & \textsc{TaPas}-small\\
    &\textsc{TaPas}-small & BERT-small & \textsc{TaPas}-small\\
    \bottomrule
\end{tabular}
\caption{Examined multi-modal bi-encoder and tri-encoder models.}
\label{table:encoders}
\end{table}

%todo let's also add a sentence about the downprojection layer either here on in the experiments section

%todo describe the training process here, e.g., batch size, epochs, what data,...
\begin{table}
\setlength{\tabcolsep}{5pt}
\centering
\begin{tabular}{lrr}
\toprule
& \textbf{Bi-encoders} & \textbf{Tri-encoders} \\
\midrule
Learning rate & 1e-5 & 1e-5\\
LR schedule & linear & linear \\
Warm-up steps & 10\% & 10\% \\
Batch size & 38 & 28 \\
Epochs & 10 & 10 \\
Optimizer & Adam  & Adam\\
\bottomrule
\end{tabular}
\caption{Hyperparameters used to train the bi-encoder and tri-encoder multi-modal retrieval models.}
\label{table:hyperparams-mm}
\end{table}
Table~\ref{table:hyperparams-mm} specifies the hyperparameters used to train the bi-encoder and tri-encoder models on the training split of the \textsc{MultiModalRetrieval} dataset described in Section~\ref{sec:datasets}. The learning objective is to create similar embeddings for relevant texts and/or tables with regard to a question. To train the models more efficiently, we make use of in-batch negatives besides each question's hard negative text or table as suggested by \citet{Karpukhin.2020} in the context of text retrieval. Given that the training samples inside a batch are randomly selected from all training examples, questions comprising a text passage as gold-label might have tables as negative labels, and vice versa. 

\section{Experiments}
\label{sec:experiments}
We compare the presented dense retrieval models to the sparse retrieval method BM25 and evaluate them based on recall@$k$ with $k \in \{10, 20, 100\} $. 
The search space to evaluate the models needs to consist of both texts and tables. 
For this purpose, 500,000 text passages are randomly sampled from \citet{Karpukhin.2020}'s preprocessed Wikipedia passages making sure that the gold passages are among these passages. 
Furthermore, besides the text passages, all the tables from \textsc{WikiSQL}, \textsc{OTT-QA}, and \textsc{NQ-Tables} are used, resulting in 656,166 tables and therefore approximately 1.2 million documents in total. 
The models are evaluated on a random sample of 1,000 questions of each dataset's test split as listed in Table~\ref{table:Datasets} and the full test split of the \textsc{MultiModalRetrieval} dataset. 

Following \citet{Karpukhin.2020}, a document retrieved for a question originating from \textsc{NQ} or \textsc{NQ-Tables} is considered a correct match if the document contains the answer string of the granular answer. 
Given that the derivation of the granular answer for questions originating from \textsc{WikiSQL} and \textsc{OTT-QA} might need further aggregation, such as summation or counting, and, therefore, the answer string does not need to be present in a relevant document, a retrieved document is only considered a correct match if it is the gold annotated table. This evaluation procedure might have the effect of incorrectly judging non-gold tables that contain the answer to a query as irrelevant. However, since we apply the same evaluation procedure for all models, the numbers should be comparable.
Table~\ref{table:MM_Results} specifies the evaluation results for BM25, all bi-encoder and all tri-encoder models. 

%\begin{sidewaystable*}[p]
\setlength{\tabcolsep}{3.5pt}
\begin{table*}
\centering
\begin{tabular}{cccrrrrrrrrr}
\toprule
\multicolumn{3}{c}{\textbf{\small Encoders}}                & \multicolumn{3}{c}{\textbf{\textsc{\small NQ}}}     & \multicolumn{3}{c}{\textbf{\textsc{\small WikiSQL}}}   & \multicolumn{3}{c}{\textbf{\textsc{\small WikiSQL}$_{\text{ctx-independent}}$}} \\
Question & Text    & Table  & R@10     & R@20   & R@100  & R@10     & R@20     & R@100   & R@10    & R@20    & R@100   \\
\midrule
\multicolumn{3}{c}{--------- BM25 ---------}                           & 53.3    & 59.8  & 73.9  & \bftab 42.1    & \bftab 47.1    & \bftab 59.5   & \bftab 61.2   & \bftab 67.2   & \bftab 81.0   \\
BERT       & \multicolumn{2}{c}{--- BERT ---}  & \bftab 70.1    & \bftab 76.0  & \bftab 84.2  & 17.5    & 22.8    & 39.6   & 30.7   & 40.9   & 59.1   \\
BERT       & \multicolumn{2}{c}{--- \textsc{TaPas} ---} & 50.0    & 57.1  & 72.7  & 6.4    & 8.9    & 20.3   & 12.3   & 17.1   & 36.7   \\
\textsc{TaPas}      & \multicolumn{2}{c}{--- \textsc{TaPas} ---} & 59.6    & 67.4  & 78.2  & 9.5    & 13.5    & 26.3   & 16.6   & 23.6   & 45.9   \\
BERT       & BERT     & BERT     & 69.1    & 75.0  & 83.5  & 20.2    & 26.8    & 43.3   & 30.8   & 38.2   & 60.7   \\
BERT       & BERT     & \textsc{TaPas}    & 59.0    & 67.1  & 77.6  & 3.5    & 5.1    & 15.2   & 7.5   & 12.0   & 30.9   \\
\textsc{TaPas}      & BERT     & \textsc{TaPas}    & 46.2    & 53.5  & 68.0  & 10.7    & 14.0    & 28.8   & 17.1   & 23.3   & 45.0   \\
\bottomrule
& & & & & & & & & & & \\
\toprule
\multicolumn{3}{c}{\textbf{\small Encoders}}  & \multicolumn{3}{c}{\textbf{\textsc{\small OTT-QA}}} & \multicolumn{3}{c}{\textbf{\textsc{\small NQ-Tables}}} & \multicolumn{3}{c}{\textbf{\textsc{\small MultiModalRetrieval}}} \\
Question & Text & Table & R@10 & R@20 & R@100  & R@10     & R@20 & R@100 & R@10 & R@20 & R@100\\
\midrule
\multicolumn{3}{c}{--------- BM25 ---------} & 40.2    & 45.6  & 58.2  & 56.6    & 65.1    & 82.8  & 50.7 & 57.0 & 71.1 \\
BERT       & \multicolumn{2}{c}{--- BERT ---}  & 72.9    & 78.0  & 89.4  & 84.9    & 91.2    & \bftab 96.7   & 55.2 & 61.8 & 73.8\\
BERT       & \multicolumn{2}{c}{--- \textsc{TaPas} ---} & 30.6    & 40.0  & 63.1  & 71.6    & 72.6    & 90.0   & 34.2 & 39.1 & 56.6\\
\textsc{TaPas}      & \multicolumn{2}{c}{--- \textsc{TaPas} ---} & 49.9    & 60.4  & 82.8  & 78.9    & 86.0    & 94.0   & 42.9 & 50.2 & 65.4\\
BERT       & BERT     & BERT     & \bftab 73.8    & \bftab 79.7  & \bftab 90.1  & \bftab 86.4    & \bftab 91.6    & \bftab 96.7 & \bftab 56.1 & \bftab 62.3 & \bftab 74.9\\
BERT       & BERT     & \textsc{TaPas}    & 25.8    & 34.3  & 59.4  & 52.3    & 62.1    & 80.7   & 29.6 & 36.1 & 52.8\\
\textsc{TaPas}      & BERT     & \textsc{TaPas}    & 50.8    & 60.9  & 79.8  & 76.9    & 82.8    & 92.9   & 40.3 & 46.9 & 62.9 \\
\bottomrule
\end{tabular}
\caption{Evaluation results of BM25 and bi-encoder and tri-encoder retrieval models on 1000 random samples of the test splits of \textsc{NQ}, \mbox{\textsc{WikiSQL}}, context-independent questions of \textsc{WikiSQL}, \textsc{OTT-QA}, and \textsc{NQ-Tables} and the full test set of our new \textsc{MultiModalRetrieval} dataset with regard to recall@10, recall@20, and recall@100.}\label{table:MM_Results}
%\end{sidewaystable*}
\end{table*}

The evaluation shows that BM25 outperforms all dense methods on both the full \textsc{WikiSQL} dataset and \textsc{WikiSQL}'s context-independent questions. 
It outperforms the best dense retrieval model on this dataset, the tri-encoder consisting of three BERT models, by 21.9 percentage points on all \textsc{Wiki\-SQL} questions and 30.4 percentage points on context-independent \textsc{WikiSQL} questions with regard to recall@10.
%todo explain that we think this is due to high lexical overlap. and that our wiki sql subset shows that with decreasing lexical overlap, bm25 performs worse up until dense retrieval outperforms sparse retrieval

Analysing the \textsc{WikiSQL} dataset in more detail shows a very high lexical overlap of questions with their accompanying table. A combination of the Jaccard coefficient and Gestalt-Pattern-Matching\footnote{We use an implementation from: \url{https://github.com/seatgeek/fuzzywuzzy\#token-set-ratio}} allows to quantify the lexical overlap of the questions with their corresponding tables without incorporating neither duplicate occurrences of the same word nor the order of the words inside the questions and the tables. This word order independence is particularly important for the analysis, given that the sparse retrieval method BM25 is order-agnostic.
%Using this combination of the Jaccard coefficient and Gestalt-Pattern-Matching\footnote{For this step, we made use of Fuzzywuzzy's implementation: \url{https://github.com/seatgeek/fuzzywuzzy\#token-set-ratio}} on the lower-cased questions and tables shows that 40.68\% of the questions overlap completely when removing the stop-words. 
Even after lower-casing questions and tables and removing stopwords in the \textsc{WikiSQL} dataset, 40.68\% of the questions lexically overlap completely with the relevant table, according to the combination of the Jaccard coefficient and Gestalt-Pattern-Matching.
This large lexical overlap explains BM25's strong performance on that dataset in Table~\ref{table:MM_Results}.
In contrast, only 0.27\% of the questions in \textsc{OTT-QA} overlap completely with their accompanying table.
For the other datasets, 15.73\% of the questions in \textsc{NQ-Tables}, 14.47\% of the questions in \textsc{NQ}, and 21.96\% of the questions in \textsc{MULTIMODALRETRIEVAL} overlap completely with their accompanying table or text.
%The average lexical overlap of questions with their corresponding table is 88.4\%.

\begin{figure*}
\centering
\begin{subfigure}{0.5\textwidth}
\centering
\includegraphics[width=0.8\textwidth]{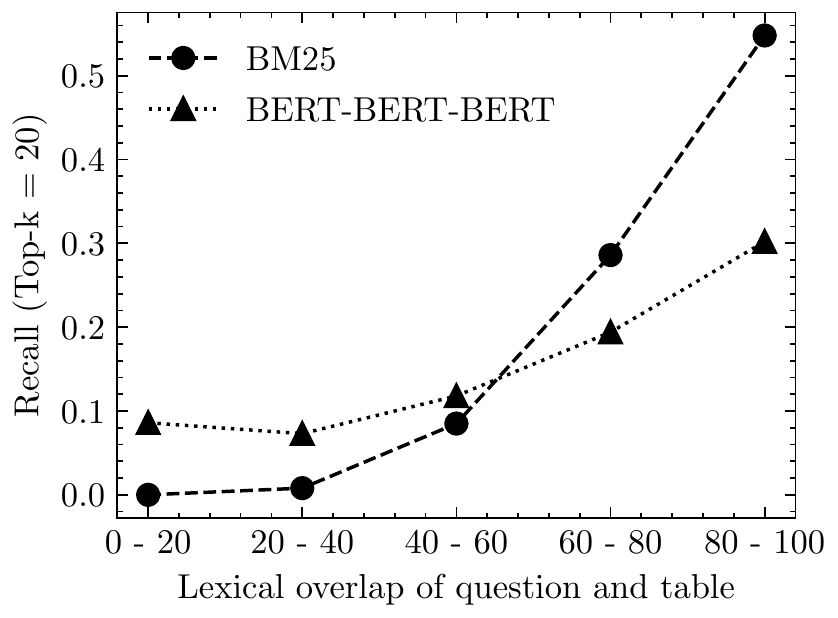}
\subcaption{\textsc{WikiSQL}}
\end{subfigure}%
\begin{subfigure}{0.5\textwidth}
\centering
\includegraphics[width=0.8\textwidth]{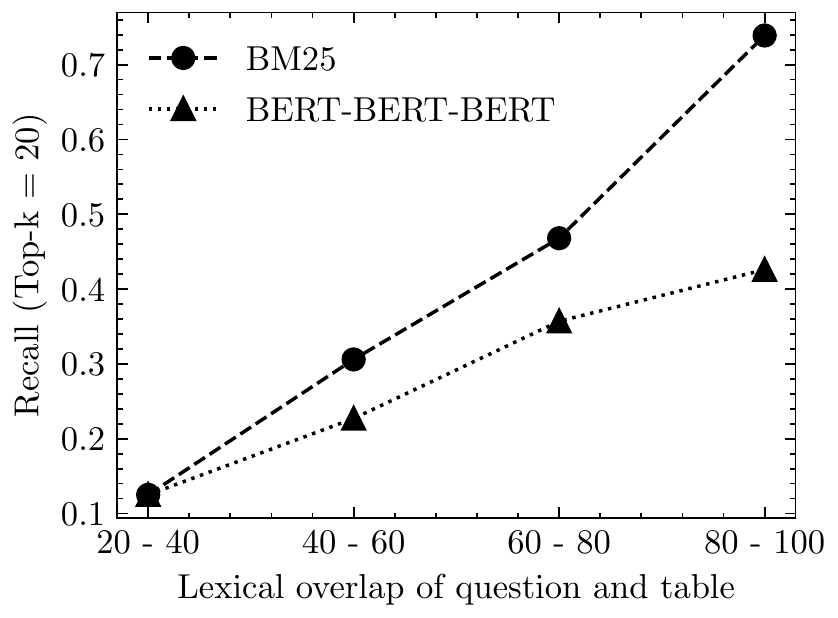}
\subcaption{\textsc{WikiSQL}$_{\text{ctx-independent}}$}
\end{subfigure}
\caption{Recall@20 of the BM25 model and the BERT-BERT-BERT tri-encoder across different percentage of lexical overlap of question and table. Performance of BM25 drops drastically if the lexical overlap is low.}
\label{fig:lexical_overlap}
\end{figure*}

To better understand how the lexical overlap of questions and accompanying tables influences BM25's performance, we split the test sets of \textsc{WikiSQL} and \textsc{WikiSQL}$_{\text{ctx-independent}}$ into subsets with different ranges of lexical overlap. As can be observed in Figure~\ref{fig:lexical_overlap}, the recall of both BM25 and dense retrieval highly correlates with lexical overlap. Furthermore, while BM25 outperforms dense retrieval for questions with high lexical overlap, it is the other way round for questions with low lexical overlap.
%The average lexical overlap of questions with their corresponding table is 88.4\%.
%we can observe that OTT-QA's distribution of lexical overlaps is much wider than WikiSQL's distribution. Only 0.27\% of OTT-QA's questions overlap completely with their accompanying table. The average lexical overlap is 66.6\%. 

On the sample of the \textsc{NQ-Tables} test set, all dense retrieval models outperform BM25, except for the tri-encoder that consists of BERT instances as question and text passage encoder and a \textsc{TaPas} instance as table encoder. 
The best performing model on this dataset is the tri-encoder consisting of three BERT encoders. This model outperforms the sparse retrieval method BM25 by 29.8 percentage points with regard to recall@10.

For the sampled questions of the \textsc{OTT-QA} development set, four out of the six dense retrieval models outperform BM25. The best performing model is again the tri-encoder model that is composed of three BERT-small encoders. This model outperforms BM25 by 33.6 percentage points with regard to recall@10. BM25 outperforms the bi-encoder consisting of a BERT model as question encoder and a \textsc{TaPas} model as text and table encoder as well as the tri-encoder consisting of two BERT models serving as question encoder and text encoder, respectively, and a \textsc{TaPas} model serving as table encoder.

When it comes to the performance on the text modality, i.e., questions deriving from \textsc{NQ} whose gold-label answer resides in a text passage, four out of the six dense retrieval models outperform BM25. For this case, the best performing model is not a tri-encoder but the bi-encoder comprising two BERT models. This model outperforms BM25 by 16.8 percentage points with regard to recall@10. However, this bi-encoder exceeds the tri-encoder consisting of three BERT encoders only slightly by one percentage point. The sparse retrieval method BM25 beats the bi-encoder consisting of a BERT model as question encoder and a \textsc{TaPas} model as text and table encoder as well as the tri-encoder consisting of two \textsc{TaPas} models serving as question encoder and table encoder, respectively, and a BERT model serving as text encoder.

In summary, the best performance on the \textsc{WikiSQL} test set is achieved by the sparse retrieval method BM25. The tri-encoder consisting of three BERT encoders shows the best performance on the remaining two tabular datasets, \textsc{OTT-QA} and \textsc{NQ-Tables}. On the \textsc{NQ} dataset, i.e., questions whose answers reside in the textual modality, the bi-encoder consisting of two BERT encoders performs best but is almost on par with the tri-encoder consisting of three BERT models.

We can conclude that, under the limited experimental conditions, in particular, on the datasets used in our study, models involving \textsc{TaPas} as question, text, and/or table encoder perform worse than models that rely only on BERT language models. \citet{Herzig.2021} show that to be able to use \textsc{TaPas} for the retrieval of tables, \textsc{TaPas} needs to be additionally pre-trained on the table retrieval task. These pre-trained table retrieval models, which are used in the bi-encoders and tri-encoders that involve one or more \textsc{TaPas} instances as encoder, are, however, pre-trained solely on the task of table retrieval and not text retrieval. Given the fact that the plain \textsc{TaPas} model cannot be adapted to retrieval from scratch but needs this special pre-training, it might be the case that, to use \textsc{TaPas} efficiently for the retrieval of both texts and tables, it needs to be pre-trained in a multi-modal setting on the retrieval of both texts and tables. Furthermore, batch size is significant for training retrieval models, as higher batch sizes make the training harder by adding more in-batch negatives. While the training of a bi-encoder does not allow a batch size higher than 38 and the training of a tri-encoder does not allow a batch size higher than 28 on a Tesla V100 GPU with 16 GB of memory, \citet{Herzig.2021} make use of a batch size of 256 for training their \textsc{TaPas}-based table retrieval models. Accordingly, it might be the case that \textsc{TaPas} is more unstable to train and requires, therefore, larger batch sizes.

%todo maybe we could add a figure to show the influence of lower/higher lexical overlap on BM25 and bi-encoder performance on WikiSQL data?

\section{Conclusion and Future Work}
\label{sec:conclusion}
This paper presented a transformer-based approach using bi-encoder and tri-encoder models for multi-modal retrieval of tables and texts.
%as one component of an open-domain QA pipeline.
With experiments on five datasets from related work and one newly created dataset, we show that the presented dense retrieval models outperform the sparse retrieval model BM25 if there is a low lexical overlap of questions and relevant tables and texts.
%
%
%Evaluation of these models on a search corpus consisting of text and tables and comparison to BM25 shows the sparse retrieval method BM25 outperforms all dense retrieval models on questions originating from the tabular dataset \textsc{WikiSQL}. However, when it comes to the remaining evaluated datasets, it can be concluded that dense retrieval models outperform the sparse retrieval model BM25. 
More specifically, the tri-encoder architecture performs better on \textsc{OTT-QA} and \textsc{NQ-Tables}, which represent the tabular modality, while the bi-encoder architecture performs slightly better on the \textsc{NQ} dataset representing the textual modality. 
We observe that the best retrieval models are those that rely only on BERT models as encoder and do not make use of \textsc{TaPas}.

%This means that models generating dense vector representations for retrieval can be, indeed, extended to a multi-modal setting to allow the joint retrieval of both text and tabular data. Additionally, the results reveal that these multi-modal dense retrieval models outperform the standard sparse retrieval method for text, BM25, in most cases. 
%Future work on this topic should, first and foremost, address the open questions of this paper. Therefore, further research should examine why the dense retrieval models perform so poorly on the \textsc{WikiSQL} dataset. Furthermore, a deep investigation on the differences between BERT and \textsc{TaPas} might explain why \textsc{TaPas} needs additional pre-training to be useful for retrieval while BERT can be easily fine-tuned for this task without additional pre-training.

%Besides, further work could try to improve the achieved results by making use of larger language models. Additionally, pre-training \textsc{TaPas} on the task of multi-modal retrieval and using higher batch sizes might make it possible to exploit \textsc{TaPas}' capabilities for multi-modal retrieval and achieve better results. Another approach to possibly improve the joint retrieval of texts and tables could be to utilize two separate retrievers, one for retrieving texts and the other for retrieving tables. The challenge with this approach is to smartly join the results of each retriever, given that the retrievers use different embedding spaces and, therefore, the scores cannot be joined naïvely. 

From an application point of view, future work could integrate the presented retrieval models as one component in a multi-modal open-domain QA pipeline and evaluate it on a real-world use case. 
Such a pipeline would facilitate information access immensely by combining valuable information from both sources rather than relying only on either texts or tables.
Another promising path for future work is to extend our approach to more modalities with transformer-based models for images, videos, or speech.
%While transformer-based models started as a technique to apply transfer learning on language-specific tasks, more and more models are released that adapt the attention mechanism characteristic for transformer-based models to non-textual data. 
%To date, there exist for images \cite{Tan.2019, Li.2020, Dosovitskiy.2021, Li.2021, Radford.2021, Touvron.2021}, videos \cite{Bertasius.2021} and speech \cite{Baevski.2020, Hsu.2021}.
These models could serve as encoders for documents of different modalities to jointly train an $n$-encoder architecture, where one encoder is tailored to the queries and the remaining $n - 1$ encoders are tailored to each of the modalities that the user would like to search on.
Last but not least, the research community would surely benefit from the creation of more multi-modal datasets to improve training and evaluation of multi-modal retrieval models and we are only making a first step in this direction with creating and releasing a dataset of tables and texts.

%\subsection{Citations}
%\citep{Gusfield:97}
%\citet
%
%\begin{table*}
%\centering
%\begin{tabular}{lll}
%\hline
%\textbf{Output} & \textbf{natbib command} & \textbf{Old ACL-style command}\\
%\hline
%\citep{Gusfield:97} & \verb|\citep| & \verb|\cite| \\
%\hline
%\end{tabular}
%\caption{\label{citation-guide}
%caption}
%\end{table*}
%Table~\ref{citation-guide}.

\section*{Acknowledgements}
We would like to thank Jonathan Herzig and Julian Eisenschlos for taking the time to discuss ideas with us and to give early feedback on experiment results.

\bibliography{anthology,custom}
\bibliographystyle{acl_natbib}

\end{document}